\documentclass[11pt]{article}

\usepackage[preprint]{acl}

\usepackage{times}
\usepackage{latexsym}

\usepackage[T1]{fontenc}

\usepackage[utf8]{inputenc}

\usepackage{microtype}

\usepackage{inconsolata}

\usepackage{graphicx}

\usepackage{amsmath}
\usepackage{amssymb}
\usepackage{booktabs}
\usepackage{bm}
\usepackage{multirow}
\usepackage{multicol}
\usepackage{tcolorbox}
\usepackage{todonotes}
\usepackage{makecell}
\usepackage{subcaption}

\title{Multilinguality of Large Language Models From a Structural Perspective}

\author{
  \textbf{Haruki Sakajo} \; \;
  \textbf{Yusuke Sakai} \; \;
  \textbf{Hidetaka Kamigaito} \; \;
  \textbf{Taro Watanabe}
\\
  Nara Institute of Science and Technology (NAIST)
\\
\texttt{sakajo.haruki.sd9@naist.ac.jp}\\
\texttt{\{sakai.yusuke.sr9, kamigaito.h, taro\}@is.naist.jp}
}

\begin{document}
\maketitle
\begin{abstract}
Large language models (LLMs) have excelled in processing multiple languages through pre- and post-training on multilingual data, even though English dominates the training data.
Prior work focusing on token representations has revealed how those LLMs process non-English text.
Although these analyses have provided insightful findings, they fail to capture a structural view, which is an inherent property of language.
In this study, we explore the multilinguality of LLMs through representational structural analysis.
Our findings reveal that low-resource languages are structurally more different from English than high- and mid-resource languages, and that language-specific post-training alters their structures while preserving inter-language relationships.
\end{abstract}

\section{Introduction}
Multilinguality of large language models (LLMs) is crucial to make LLMs helpful for everyone.
While English is the dominant language in LLM pre-training data, LLMs exhibit multilingual capabilities~\cite{winata-etal-2021-language, gemma3, qwen3}.
LLMs have improved performance across multiple languages by learning languages through pre-training on multilingual corpora~\cite{gemma3, qwen3} and continuous pre-training~\cite{fujii2024continual}.
This multilingualism of these LLMs has attracted computational linguists, and prior work has addressed fundamental questions about how these LLMs handle diverse languages: how linguistic knowledge is internally represented~\cite{brinkmann-etal-2025-large}, whether there are cross-linguistic differences~\cite{chang-etal-2022-geometry, wu-etal-2024-representational, zhang2025the}, and whether a pivot language underlies multilingual processing~\cite{wendler-etal-2024-llamas, zhong-etal-2025-language}.

Nonetheless, prior analyses have been conducted predominantly at the token (or subword) level~\cite{wendler-etal-2024-llamas, brinkmann-etal-2025-large, zhong-etal-2025-language}, although languages exhibit structures~\cite{Bybee_2010} and diverse morphosyntactic characteristics~\cite{Song2018-ug}, and tokenizers do not account for these properties.
Consequently, the semantic granularity of the minimal units processed by LLMs varies substantially across languages due to differences in tokenizers.
This limits the token-level analysis to understand how LLMs process and represent multiple languages.
This motivates an analysis that moves beyond the token level and adopts a holistic perspective that considers the entire input structure.

\begin{figure}
    \centering
    \includegraphics[width=0.92\linewidth]{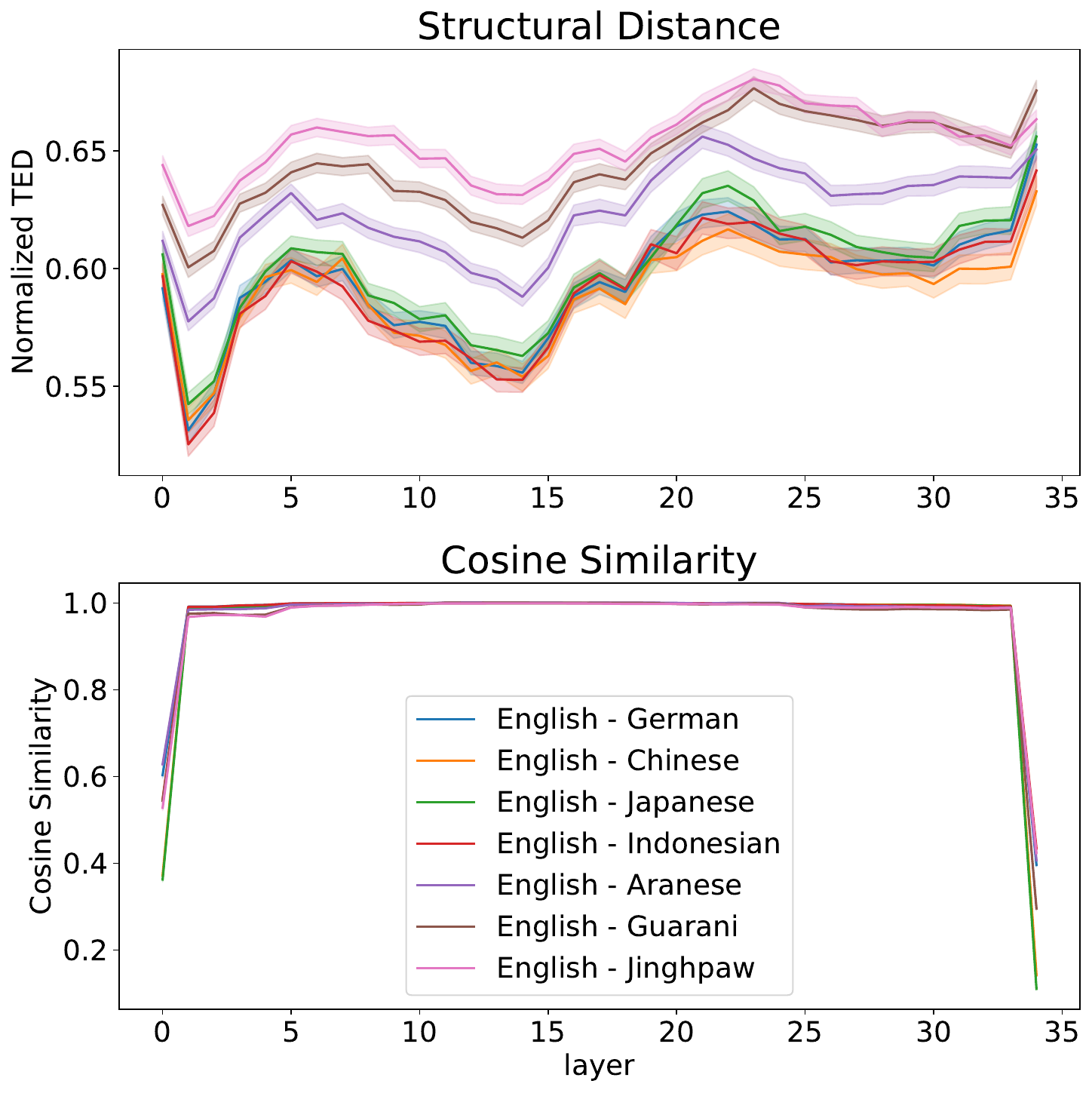}
    \caption{Representational structure distance and cosine similarity between English and other languages.}
    \label{fig:structures_cos_sim}
\end{figure}

To address this limitation, we analyze multilingual representations from a structural view using \textsc{StructLens}~\cite{structlens}, a method that derives tree structures from pairwise L2 distances between token representations.
This framework enables comparison of entire languages from a tree-structure perspective in representation space.
We measure structural distance in representation space across diverse languages, including low-resource languages, on monolingual and multilingual pre-trained models and language-specific post-trained models.
The resulting trees and their distance provide the organization of multilingual representations in the LLMs' representation space, the differences among languages, and the influence of language-specific post-training.

Our experiments reveal that a monolingual, English-centric model exhibits greater structural distance between English and other languages, whereas multilingual models display more uniform internal structures across high- and mid-resource languages while diverging in low-resource languages, despite showing high cross-lingual cosine similarity~(see Figure~\ref{fig:structures_cos_sim}).
The results also show that a language that is syntactically and typologically similar to English is relatively close to English.
Furthermore, language-specific post-training differentiates the internal structures of the target language from those of other languages, while preserving the structural distance across languages.
We also demonstrate that representational structure distance captures inter-token relationships that conventional cosine similarity fails to reflect. 
Our findings reveal that a representational structure perspective yields distinct measures of language similarity based on the multilinguality of LLMs, their training data, and the representational characteristics of language-specific post-training.

\begin{figure*}[t]
    \centering
    \includegraphics[width=\linewidth]{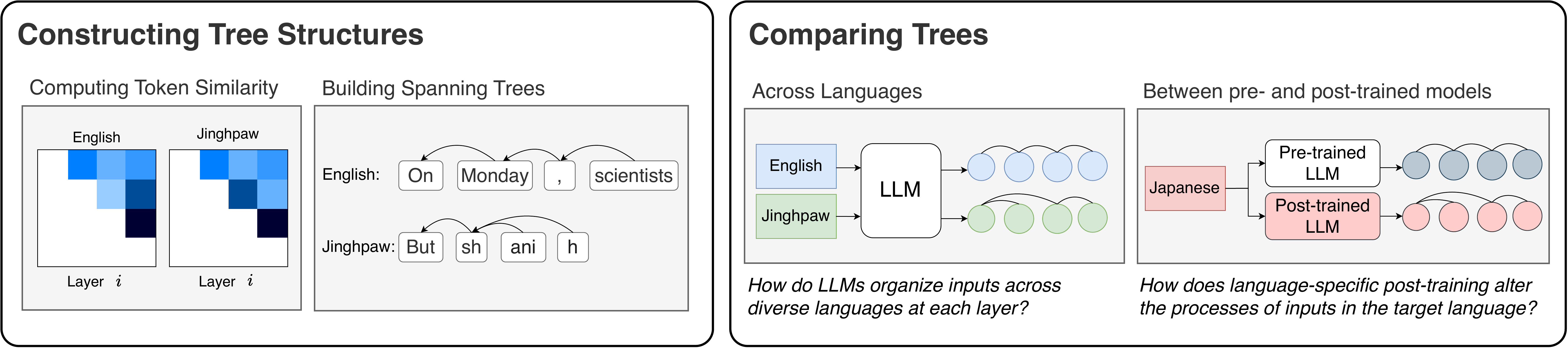}
    \caption{Overview of our research. We construct tree structures that summarize inter-token relationships in representation space and compare them across languages within a model and between pre- and post-trained models. This comparison reveals the organization of input tokens across languages and the influence of language-specific post-training.}
    \label{fig:overview}
\end{figure*}

\section{Background and Related Work}

\subsection{Representations in Residual Streams}
The pre-layer normalization~\cite{xiong-etal-onlayer} variant of Transformer~\cite{vaswaniAttentionAllYou2017} refines internal representations incrementally through residual connections, which is also known as a \emph{residual stream}~\cite{mathematicalframework}.
Residual streams reflect models' hidden states, and prior research focusing on them has demonstrated mechanisms of LLMs~\cite{mathematicalframework, kamigaito2025diversitytransformerlayersaspect}.
To interpret those states, several tools have been developed.
For example, Logit lens~\cite{logitlens} applies the language model head of LLMs for hidden states at each layer and interprets them as symbols.
Another tool, \textsc{StructLens}~\cite{structlens}, constructs spanning trees based on the L2 distance between hidden states at each layer, providing a holistic view of their internal structures in representation space.
In the remainder of this section, we refer to hidden states in the residual streams as representations.

\subsection{Language Adaptation for Multilinguality}
The training data for most LLMs is predominantly English, and their performance in non-English languages is limited compared to English~\cite{xuan-etal-2025-mmlu}. 
To fill this gap, substantial efforts have been made to transfer or extend those models' capabilities to other languages by adapting the models' vocabularies for target languages and continuous pre-training~\cite{minixhofer-etal-2022-wechsel, minixhofer2024zeroshot, yamaguchi-etal-2024-empirical, yamaguchi-etal-2024-how, remy2024transtokenization, fujii2024continual}.

Continuous pre-training on target language data is an approach to achieving language adaptation, while keeping the performance in the source language~\cite{wang-etal-2020-extending, fujii2024continual}.
For example, \citet{fujii2024continual} develops language-specific language models by vocabulary expansion and continuous pre-training on English-centric models, achieving advanced performance in both English and the target language.
We refer to continuous pre-training as language-specific post-training for convenience.

\subsection{Multilinguality of Large Language Models}
LLMs obtain multilingual capabilities through pre-training on multilingual data~\cite{gemma3, qwen3} and language-specific continuous pre-training~\cite{fujii2024continual}.
Previous work investigates how the multilingual LLMs internally represent and process different languages.
\citet{chang-etal-2022-geometry} analyzes the geometry of multilingual representations, showing that languages occupy similar linear subspaces.
\citet{wendler-etal-2024-llamas} track how representations evolve across layers when English-centric models process non-English input using Logit lens, and find that the latent concept in the models is biased toward English.
In this line of work, \citet{zhong-etal-2025-language} demonstrate that LLMs trained on balanced multiple language resources rely on those languages.
\citet{zhang2025the} discover that models employ the same circuits to process the same syntactic features across languages.
\citet{brinkmann-etal-2025-large} train Sparse Autoencoders and find that models share features for grammatical categories.

These studies have revealed that LLMs possess language-agnostic and language-specific states through analysis of token representations and circuits.
While prior work has focused on token-level representations and has revealed the underlying multilingual processes and the geometry of multilingual representations, this approach may fail to capture inherent differences in representational structure.
In this work, we aim to observe representations of entire input across languages from a structural perspective.

\section{Task Formulation}
We ask: (1) \emph{how do LLMs organize inputs across diverse languages at each layer?} (2) \emph{how does language-specific post-training alter the processes of inputs in the target language?}
We examine these research questions by investigating the inter-token relationships formed in the models' representation space.
To answer the first question, we compute the distance between languages in the LLM representation space using parallel corpora.
For the second question, we measure the distance between the tree structures of pre-trained and language-specific post-trained models.

We compute the structural distance of token representation organization in the LLMs' representation space between languages for each layer by constructing spanning trees for each language using \textsc{StructLens}~\cite{structlens} and measuring the distance between them.
\textsc{StructLens} summarizes inter-token relationships in representation space, and we can analyze the differences between internal structures of languages by comparing the resulting spanning trees, as shown in Figure~\ref{fig:overview}.

\subsection{Spanning Tree Construction}
We construct a spanning tree from representations of each layer using \textsc{StructLens}.
\textsc{StructLens} constructs maximum spanning trees based on inter-token similarity in representation space at each layer, summarizing the token representation organization.

Formally, given a token sequence of length $n$, let $d$ denote the hidden dimension of a model and  $\bm{h}^{(\ell)}_i \in \mathbb{R}^{d}$ denote the $i$-th token's representation in the residual streams of the $\ell$-th Transformer block, which are representation immediately after that block.
\textsc{StructLens} constructs maximum spanning trees for each layer using the adjacency matrix $\bm{A} \in \mathbb{R}^{n \times n}$, which represents inter-token similarity.
For $i, j \in \{1, 2, \dots, n\}$, the element of $\bm{A}$ is defined as:
\begin{equation}
\label{eq:adjacency}
    A_{i, j} = 
    \begin{cases}
        \dfrac{1}{1 + \|{\bm{h}}^{(\ell)}_i - \bm{h}^{(\ell)}_j\|} & \text{if } i < j, \\
        0 & \text{otherwise}.
    \end{cases}
\end{equation}
Following \citet{structlens}, we build the maximum spanning trees using $\bm{A}$ and the algorithm introduced by \citet{Tarjan1977} running in $O(n^2)$ time.

The resulting spanning trees summarize inter-token relationships within the representation space of each Transformer block in an LLM, providing a holistic view of each Transformer layer.
Therefore, we can answer the research questions by analyzing the spanning trees across layers and across models.

\subsection{Measuring Structural Distance}
To measure the structural differences between two trees, we compute the Tree Edit Distance~(TED)~\cite{zhangshasha}, normalized by the larger input lengths of the two trees for fair comparison, since different languages or tokenizers can yield different lengths of token sequences.
TED measures the minimum cost required to transform one tree into another using three operations: insertion, deletion, and relabeling.
Higher TED indicates lower similarity, whereas lower TED indicates higher similarity.
We set the relabeling cost to 0 when computing TED to focus on structural differences rather than lexical mismatches.
For computational efficiency, we use AP-TED~\cite{PAWLIK2016157}, an optimized variant of TED.

\begin{table*}[t]
    \centering
  \resizebox{\textwidth}{!}{
  \begin{tabular}{@{}lcccccccc@{}}
    \toprule
    Feature & \multicolumn{1}{c}{English} & \multicolumn{1}{c}{German} & \multicolumn{1}{c}{Japanese} & \multicolumn{1}{c}{Chinese} & \multicolumn{1}{c}{Indonesian} & \multicolumn{1}{c}{Aranese} & \multicolumn{1}{c}{Guarani} & \multicolumn{1}{c}{Jinghpaw} \\
    \midrule
    Language Family & Indo-European & Indo-European & Japanese & Sino-Tibetan & Austronesian & Indo-European & Tupian & Sino-Tibetan \\
    Script & Latin & Latin & Kanji/Kana & Hanzi & Latin & Latin & Latin & Latin \\
    \makecell[l]{Nominative-Accusative Alignment} & \checkmark & \checkmark & \checkmark &  & \checkmark & \checkmark & \checkmark & \checkmark \\
    Word Order & SVO & (Mix) & SOV & SVO & SVO & SVO & SVO & SOV \\
    Adposition & Prepositions & Prepositions & Postpositions & -- & Prepositions & Prepositions & Postpositions & Postpositions \\
    \makecell[l]{Order of Relative Clause and Noun} & NRel & NRel & RelN & RelN & NRel & NRel & NRel & RelN \\
    \bottomrule
    \end{tabular}
    }
    \caption{Example of features of each language. In the order of relative clause and noun, NRel denotes the order of noun-relative clause, and RelN denotes the order of relative clause-noun.}
    \label{tab:language-features}
\end{table*}

\section{Experimental Settings}
\label{sec:experimental-settings}
\subsection{Models, Languages, and Datasets}
\paragraph{Models.}
We use Gemma3 4B PT~\cite{gemma3} and  Qwen3 8B Base~\cite{qwen3}, which are multilingual pre-trained models, Olmo3 7B Base~\cite{olmo3}, which is a monolingual pre-trained model, and Qwen3 Swallow 8B CPT~\cite{fujii2024continual}, which is a continuous post-trained model on Japanese data.
We also use Qwen3 8B, a post-trained model of Qwen3 8B Base, for comparison with Qwen3 Swallow 8B CPT, as both are derived from Qwen3 8B Base.

\paragraph{Languages.}
We use English, German, Chinese, Japanese, and Indonesian as high- or mid-resource languages, and Aranese, Guarani, and Jinghpaw as low-resource languages.
Table~\ref{tab:language-features} shows examples of linguistic features based on WALS~\cite{wals}, \citet{grammar-guarani}, and \citet{aranese}.
We use diverse languages in terms of language family and typological and syntactic features.
This diverse language set enables the assessment of whether and how linguistic distance determines the language process in LLMs.
For cross-lingual structural comparison, we compare English and Chinese with other languages using corresponding texts, since English is the dominant language in the training data and Qwen3 exhibits advanced performance on Chinese~\cite{qwen3}.
To investigate the effect of language-specific post-training, we focus on English, Chinese, and Japanese.

\paragraph{Datasets.}
We utilize translation sentence pairs from FLORES+~\cite{nllb-24} and corresponding Jinghpaw data~\cite{kurabe-2013-kachin-folktales, kurabe-2017-kachin-culture-history, kurabe-2020-jinghpaw-dictionary, kurabe-2020-jinghpaw-grammar, kurabe-2020-jinghpaw-reader, taguchi-etal-2025-languages}.
We use the development set of 997 samples.

\paragraph{Prompts and implementations.}
We report the detailed experimental settings, including prompts and implementation, in Appendix~\ref{sec:detail-experimental-settings}.

\begin{table*}[t]
  \centering
  \resizebox{\textwidth}{!}{
  \begin{tabular}{lcccccccc}
    \toprule
    Model & \multicolumn{1}{c}{English} & \multicolumn{1}{c}{German} & \multicolumn{1}{c}{Japanese} & \multicolumn{1}{c}{Chinese} & \multicolumn{1}{c}{Indonesian} & \multicolumn{1}{c}{Aranese} & \multicolumn{1}{c}{Guarani} & \multicolumn{1}{c}{Jinghpaw} \\
    \midrule
    Gemma3 4B PT & $41.23_{\pm 52.20}$ & $29.10_{\pm 42.86}$ & $49.71_{\pm 52.02}$ & $131.85_{\pm 751.11}$ & $57.81_{\pm 89.31}$ & $442.45_{\pm 2978.25}$ & $360.08_{\pm 695.32}$ & $860.28_{\pm 739.95}$ \\
    Qwen3 8B Base & $31.62_{\pm 36.02}$ & $13.35_{\pm 11.58}$ & $17.77_{\pm 12.67}$ & $50.97_{\pm 64.19}$ & $14.60_{\pm 11.35}$ & $289.56_{\pm 831.69}$ & $434.38_{\pm 381.41}$ & $297.32_{\pm 179.87}$ \\
    Olmo3 7B Base & $31.08_{\pm 38.59}$ & $19.76_{\pm 16.04}$ & $10.46_{\pm 5.25}$ & $12.21_{\pm 8.16}$ & $33.92_{\pm 32.56}$ & $588.81_{\pm 1282.25}$ & $765.71_{\pm 579.38}$ & $444.25_{\pm 271.71}$ \\
    Qwen3-8B & $53.04_{\pm 108.56}$ & $19.56_{\pm 21.06}$ & $26.78_{\pm 22.19}$ & $81.29_{\pm 129.33}$ & $21.30_{\pm 22.69}$ & $561.61_{\pm 3422.74}$ & $695.72_{\pm 736.64}$ & $462.71_{\pm 315.02}$ \\
    Qwen3 Swallow 8B CPT v0.2 & $28.24_{\pm 28.59}$ & $14.81_{\pm 15.54}$ & $14.14_{\pm 9.44}$ & $91.03_{\pm 161.77}$ & $16.39_{\pm 13.98}$ & $305.87_{\pm 717.26}$ & $503.91_{\pm 496.35}$ & $325.03_{\pm 196.47}$ \\
    \bottomrule
  \end{tabular}
  }
  \caption{Perplexity comparison across models and subsets. Subscripts denote standard deviation. Perplexity is normalized by input length for fair comparison across different input lengths.}
  \label{tab:perplexity}
\end{table*}

\begin{table*}[t]
  \centering
  \small
  \begin{tabular}{lccccccc}
    \toprule
    Model & German & Japanese & Chinese & Indonesian & Aranese & Guarani & Jinghpaw \\
    \midrule
    Gemma3 4B PT & $35.36$ & $21.31$ & $22.82$ & $33.80$ & $16.39$ & $3.88$ & $1.22$ \\
    Qwen3 8B Base & $36.90$ & $22.25$ & $24.99$ & $36.07$ & $19.48$ & $5.41$ & $3.60$ \\
    Olmo3 7B Base & $35.42$ & $19.44$ & $21.87$ & $31.70$ & $13.12$ & $2.40$ & $1.99$ \\
    \bottomrule
  \end{tabular}
  \caption{MT evaluation results of BLEU. The higher values are better.}
  \label{tab:mt_bleu}
\end{table*}

\begin{figure*}[t!]
    \centering
    \includegraphics[width=0.95\linewidth]{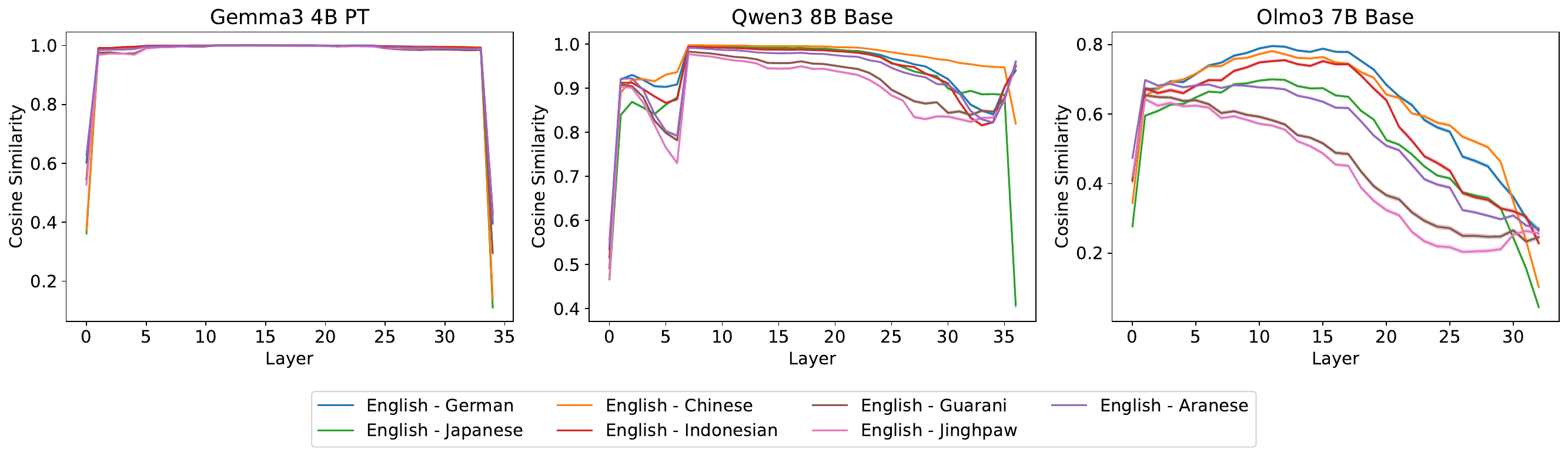}
    \caption{Cosine similarity on pre-trained models.}
    \label{fig:cos_sim}
\end{figure*}

\begin{figure}
    \centering
    \includegraphics[width=\linewidth]{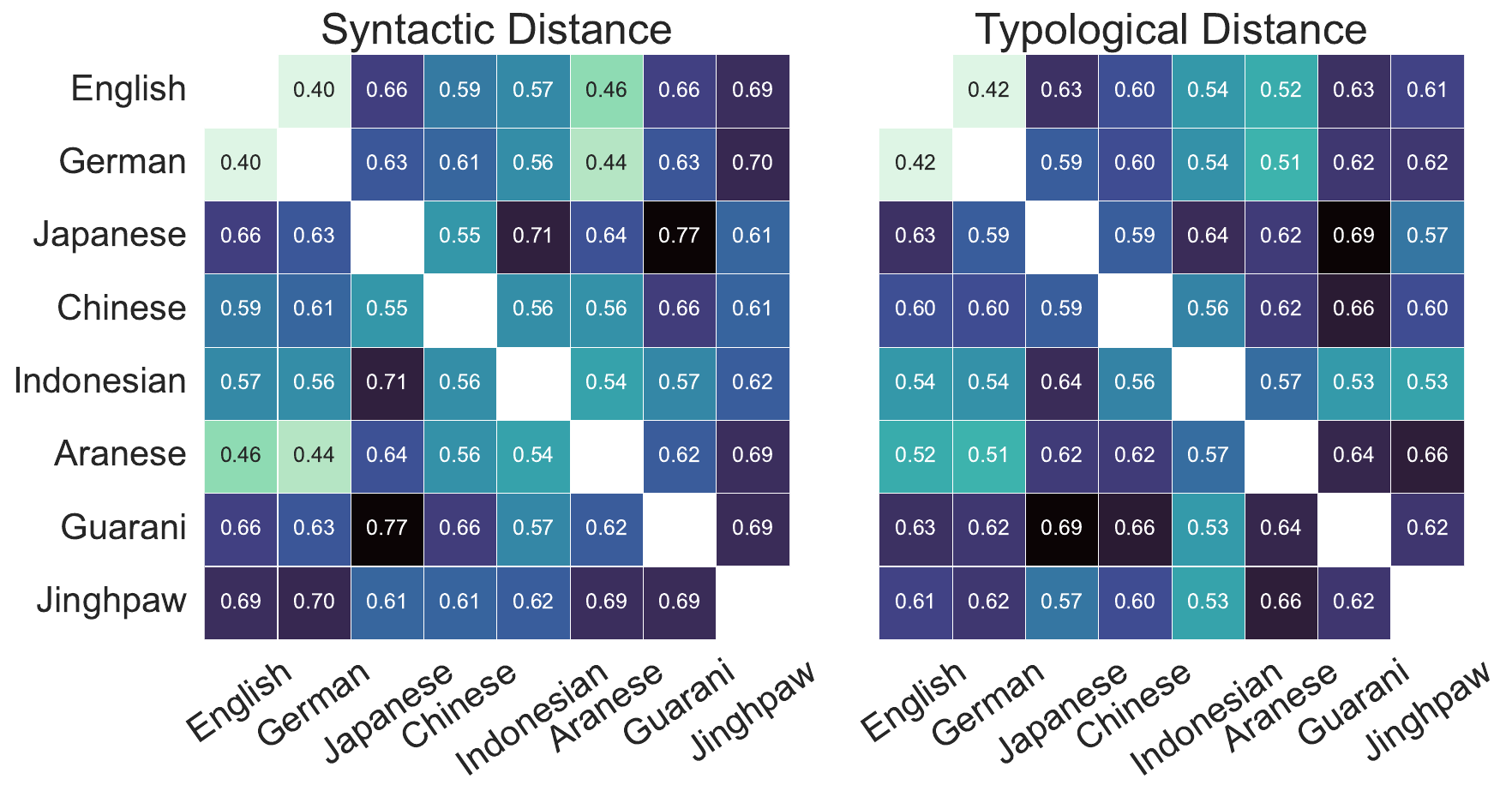}
    \caption{Syntactic and typological distance. Brighter colors represent a smaller distance, and darker colors represent a larger distance.}
    \label{fig:linguistic-distance}
\end{figure}

\subsection{Non-Structural Metrics}
\label{sec:non-structural-metrics}
To further explore what the structural metric measures, we leverage non-structural metrics.
We measure perplexity, machine translation performance, and cosine similarity.
We also measure L2 distance. See Appendix~\ref{sec:detail-non-structural-metrics-settings} for details.

\paragraph{Perplexity.}
To assess the model's familiarity with each language, we measure its perplexity.
Perplexity is computed for each sample from FLORES+ and the Jinghpaw variant in the respective language, and then normalized by the input length to ensure a fair comparison across languages and tokenizers.

\paragraph{Machine translation.}
To investigate how much LLMs can relate one language to another, we perform machine translation~(MT) experiments for each model.
In this experiment, we use non-English languages as source languages and English as a target language.
We evaluate the translated text using BLEU~\cite{papineni-etal-2002-bleu,post-2018-call}, chrF~\cite{popovic-2015-chrf}, and TER~\cite{snover-etal-2006-study}.
We truncate the generated text by period to extract the first sentence for evaluation.

\paragraph{Cosine similarity}
To measure how close semantic representations of non-English languages are to English, we compute the sentence-wise cosine similarity for each layer.
Formally, let $\bm{h}^{(\ell)}_{src}$ and $\bm{h}^{(\ell)}_{tgt}$ denote the $\ell$-th layer's representations of the source language's and target language's inputs, respectively. The cosine similarity is computed by:
\begin{equation}
\label{eq:cos-sim}
\begin{split}
    &\frac{\bar{\bm{h}}^{{(\ell)}^\top}_{src} \bar{\bm{h}}^{(\ell)}_{tgt}}{\| \bar{\bm{h}}^{(\ell)}_{src}\| \| \bar{\bm{h}}^{(\ell)}_{tgt} \|},\\
    &\quad\text{where}\ \bar{\bm{h}}^{(\ell)} = \frac{1}{|\bm{h}^{(\ell)}|}\sum\bm{h}^{(\ell)}_i.
\end{split}
\end{equation}

\subsection{Syntactic and Typological Metrics}
\label{sec:typological-metric}
To compute linguistic distance between languages, we utilize URIEL+~\cite{khan-etal-2025-uriel}, a knowledge base that stores linguistic features as vector representations.
URIEL+ computes the angle between the feature vectors of two languages, following \citet{toossi-etal-2024-reproducibility}, thereby providing the language distance with respect to syntax and typology.
We measure both syntactic and typological distance between languages.

Formally, let $m$ denote the feature dimension, and $\bm{u} \in \mathbb{R}^m$ and $\bm{v} \in \mathbb{R}^m$ denote feature vectors of two languages.
The distance of these languages is defined as:
\begin{equation}
    \dfrac{1}{\pi}\arccos{\dfrac{\bm{u}^\top \bm{v}}{\|\bm{u}\|\|\bm{v}\|}}.
\end{equation}
Following \citet{khan-etal-2025-uriel}, we use \textsc{Soft-Impute}~\cite{softimpute} to impute missing features before computing distance.

\begin{figure*}[t]
    \centering
    \includegraphics[width=\linewidth]{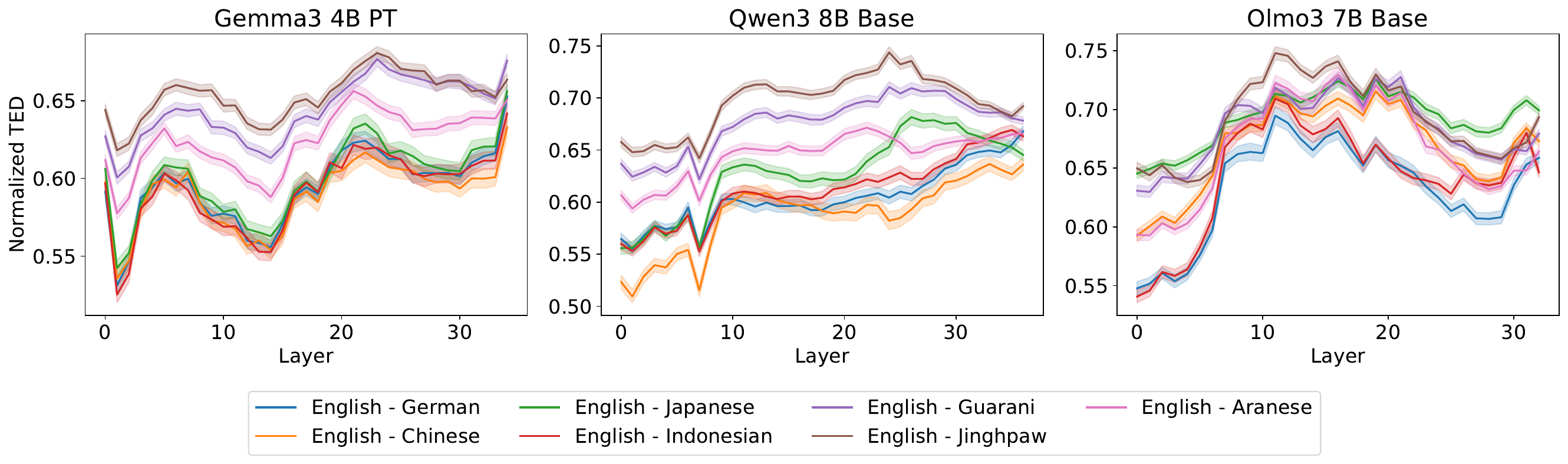}
    \caption{TED between English and other languages. The error bars indicate the 95 \% confidence interval.}
    \label{fig:ted_eng_tgt}
\end{figure*}

\begin{figure*}
    \centering
    \includegraphics[width=\linewidth]{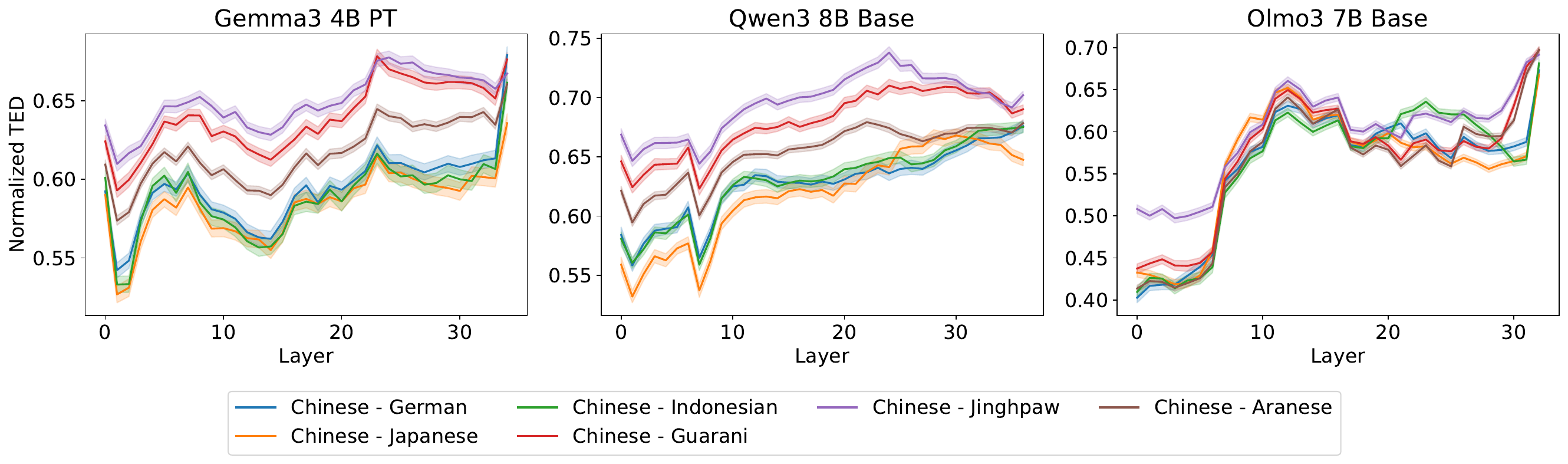}
    \caption{TED between Chinese and other languages. The error bars indicate the 95 \% confidence interval.}
    \label{fig:ted_zh_tgt}
\end{figure*}

\begin{figure*}[t]
    \centering
    \includegraphics[width=0.9\linewidth]{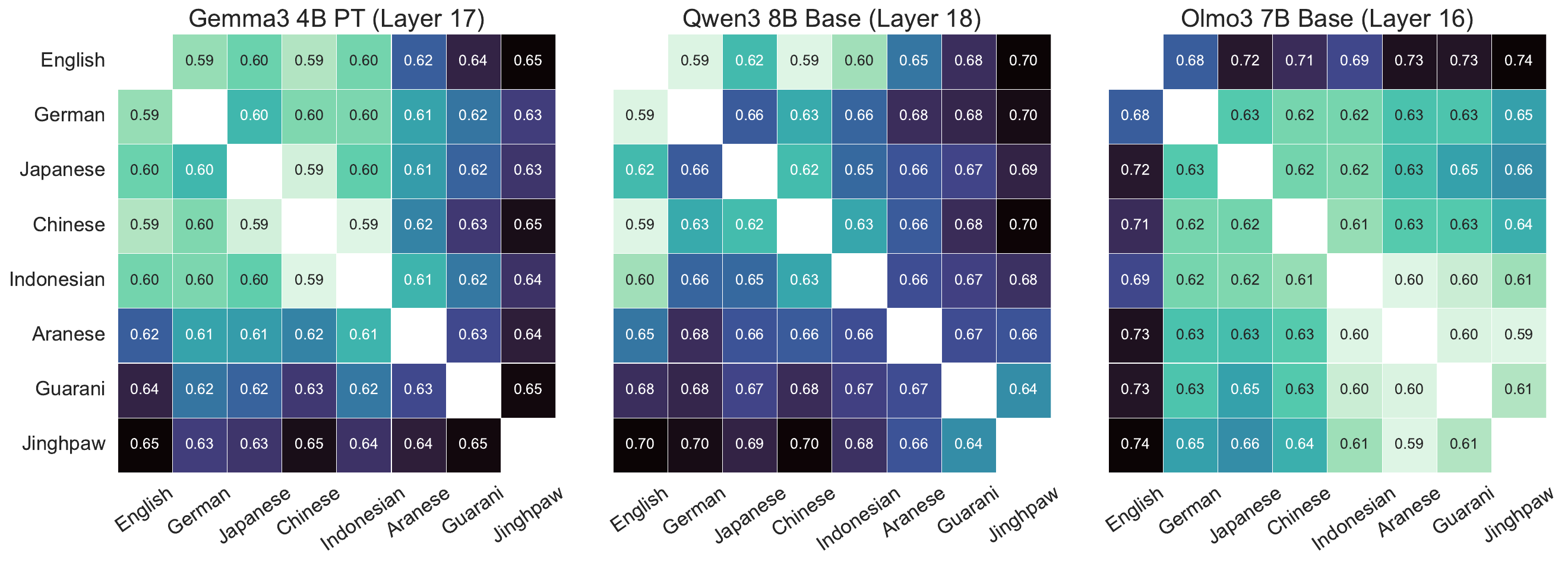}
    \caption{Heatmaps of TED in a middle layer for each language pair. Brighter colors represent a smaller distance, and darker colors represent a larger distance.}
    \label{fig:ted_heatmaps}
\end{figure*}

\begin{figure*}[t]
    \centering
    \includegraphics[width=\linewidth]{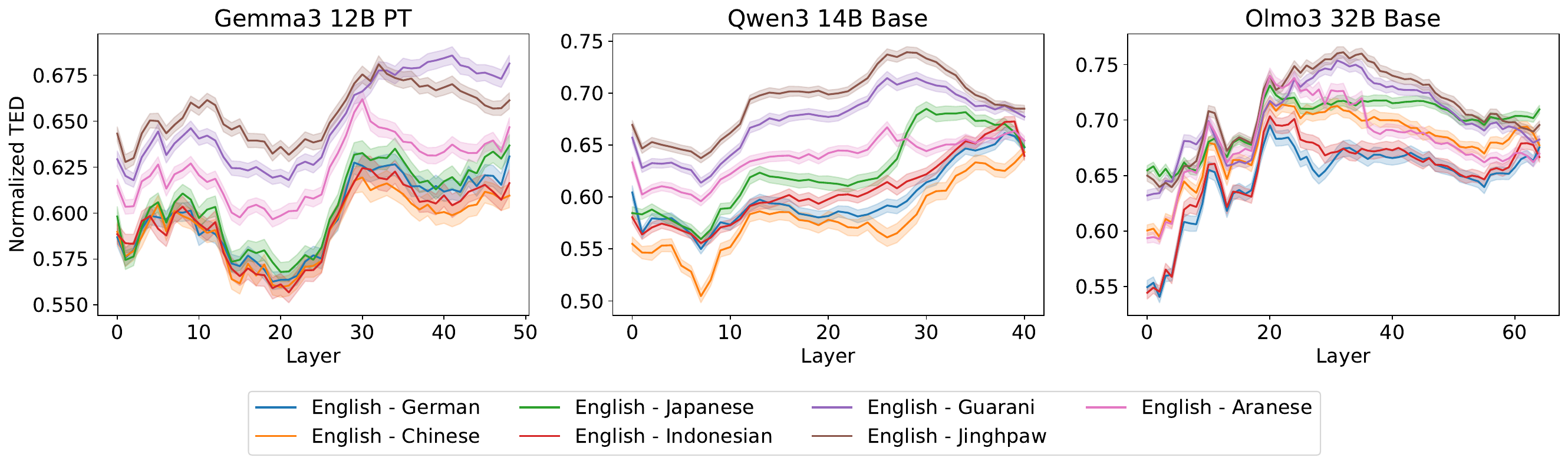}
    \caption{TED between English and other languages for larger models. The error bars indicate the 95 \% confidence interval.}
    \label{fig:ted_eng_tgt_medium}
\end{figure*}

\begin{figure*}
    \centering
    \begin{subfigure}[b]{0.6\linewidth}
        \centering
        \includegraphics[width=\linewidth]{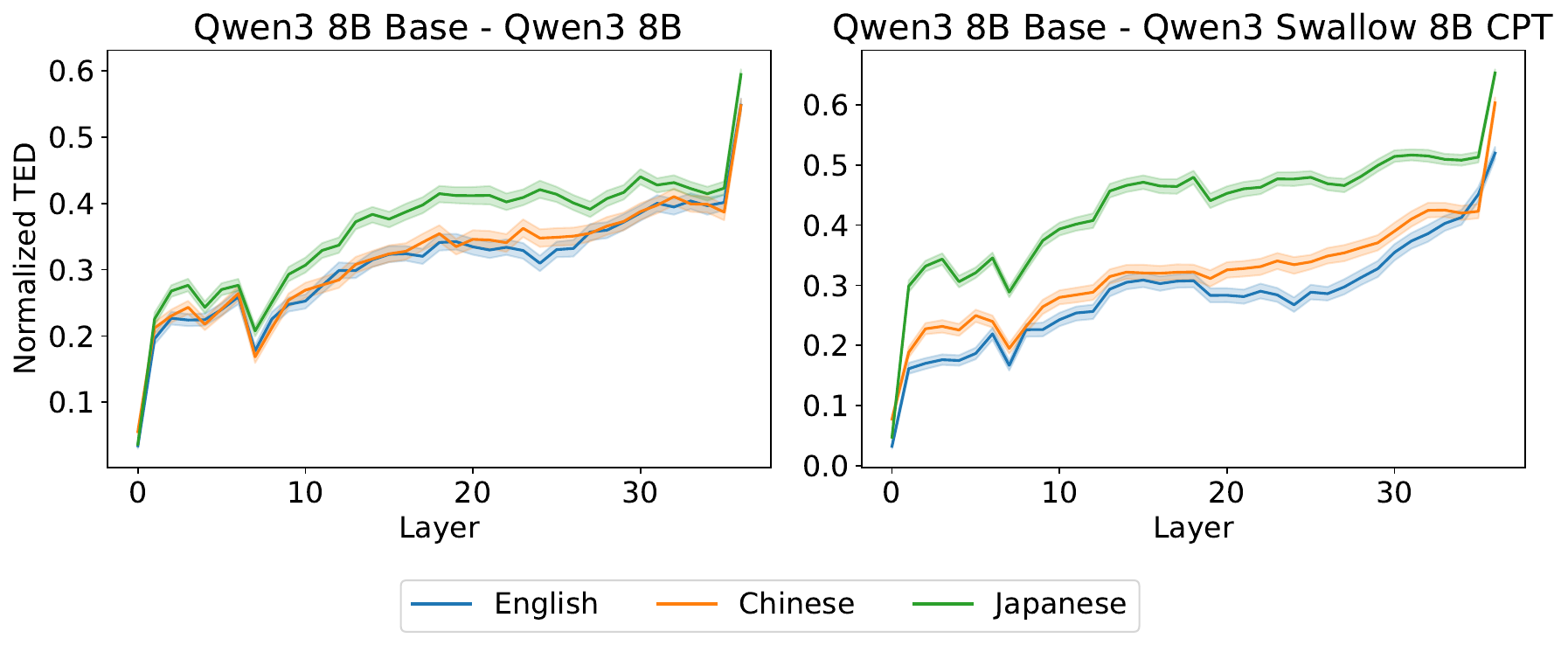}
        \caption{TED between pre- and post-trained models.}
        \label{fig:ted_post_train}
    \end{subfigure}
    \begin{subfigure}[b]{0.35\linewidth}
        \centering
        \includegraphics[width=\linewidth]{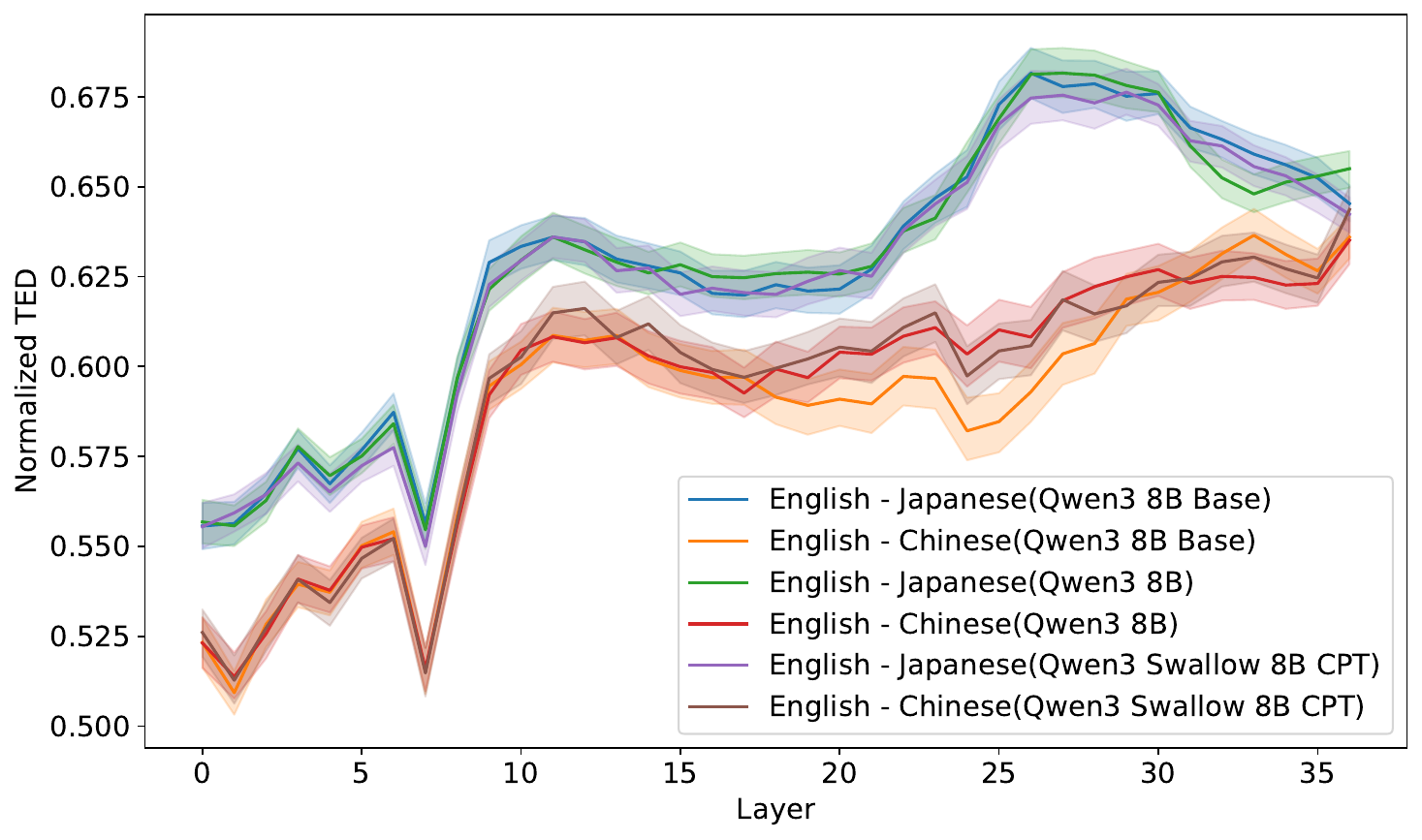}
        \caption{TED between English and \{Chinese, Japanese\} for pre- and post-trained models.}
        \label{fig:ted_eng_tgt_post_train}
    \end{subfigure}
    \caption{TED for post-trained models. The error bars indicate the 95 \% confidence interval}
    \label{fig:ted_post_train_all}
\end{figure*}

\begin{figure*}[t]
    \centering
    \includegraphics[width=0.85\linewidth]{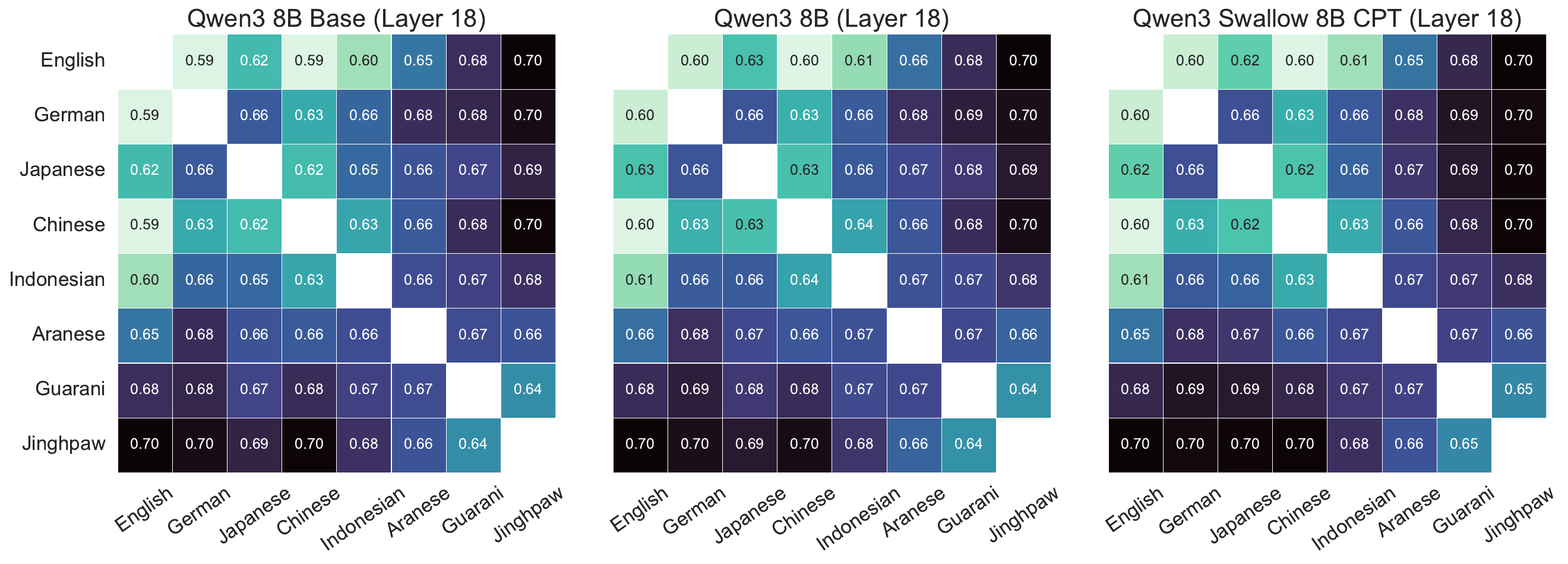}
    \caption{Heatmaps of TED in a middle layer of pre- and post-trained models for each language pair.}
    \label{fig:ted_heatmap_posttrained}
\end{figure*}

\section{Results and Discussions}

\subsection{Non-Structural Metrics}

\paragraph{Perplexity.}
Table~\ref{tab:perplexity} reports the mean perplexity for each language.
These results indicate that the low-resource languages Aranese, Guarani, and Jinghpaw are unfamiliar to the models.

\paragraph{Machine Translation.}
Table~\ref{tab:mt_bleu} shows the machine translation results measured by BLEU.
The results for other MT metrics are reported in Appendix~\ref{sec:detail-non-structural-metrics}, which show patterns similar to BLEU.
MT results indicate that Guarani and Jinghpaw are unfamiliar languages for the models.
While Aranese exhibits higher perplexity, models show better MT performance for this language than for the other low-resource languages.

\paragraph{Cosine similarity.}
Figure~\ref{fig:cos_sim} shows cosine similarity for pre-trained models.
Cosine similarity between English and other languages at intermediate layers of multilingual pre-trained models, Gemma3 4B PT and Qwen3 8B Base, is close to 0.9 across all languages.
A monolingual model, Olmo3 7B Base, exhibits lower cosine similarity for low-resource languages, whereas it shows relatively high cosine similarity for high-/mid-resource languages.
Results on post-trained models are provided in Appendix~\ref{sec:detail-experimental-results}, indicating that post-training has less impact on cosine similarity.

\paragraph{Syntactic and typological distance.}
Figure~\ref{fig:linguistic-distance} shows the syntactic and typological (featural) distance of each language pair.
German and Aranese are relatively close to English, while other languages are more distant from English, which differs from other metrics.
As for closeness, Japanese and Indonesian are relatively close to Chinese.

\subsection{Structural Metric}

Figure~\ref{fig:ted_eng_tgt} shows that there are two clusters of languages, high-/mid-resource languages and low-resource languages, based on TED in Gemma3 4B PT and Qwen3 8B Base, while Olmo3 7B Base shows equivalent results across languages, which are different from MT results and cosine similarity.
While linguistic typological features, e.g., word order and morphological features (see Table~\ref{tab:language-features}), can affect TED between languages, even languages with diverse typological features and scripts (i.e., German, Chinese, Japanese, and Indonesian) show a close distance to English.
Given that Gemma3 4B PT and Qwen3 8B Base are multilingual models, and Olmo3 7B Base is a monolingual model, the results indicate that pre-training data enable models to distinguish the structural distance.
Figure~\ref{fig:ted_zh_tgt} reports TED between Chinese and other languages, yielding similar results with Figure~\ref{fig:ted_eng_tgt}.
We also investigate larger models within each model family, as shown in Figure~\ref{fig:ted_eng_tgt_medium}, which exhibit similar patterns.
We further visualize TED for each language pair in the middle layer of Figure~\ref{fig:ted_heatmaps} as a case study, showing that Gemma3 4B PT exhibits a cluster among high- and mid-resource languages, and that Olmo3 7B Base exhibits distinct English and non-English patterns.
Appendix~\ref{appendix:structural-distance} provides the TED matrices for the low and high layers.

Figure~\ref{fig:ted_post_train} shows the TED between pre- and post-trained models for English, Chinese, and Japanese, and Figure~\ref{fig:ted_eng_tgt_post_train} shows TED between English and {Chinese, Japanese} for pre- and post-training models.
We also show TED for each language pair in Figure~\ref{fig:ted_heatmap_posttrained}.
These figures indicate that the Japanese-specific post-trained model exhibits higher TED in Japanese, while TED between English and Chinese and between English and Japanese is consistent with that of the pre-trained model.

\subsection{Difference between Cosine Similarity and Structural Distance}
\label{sec:diff-cos-sim-ted}
The cosine similarity measures the difference between the centroids of each language's inputs, and the mean representations fail to capture relationships among token representations.
To formalize this difficulty, we decompose the residual streams into the mean representation and the difference from it.
Let $\bm\delta_i^{(\ell)}$ denote the difference representation from $\bar{\bm{h}}^{(\ell)}$, which is defined in Equation~\ref{eq:cos-sim}.
We decompose the representation $\bm{h}_i^{(\ell)}$ as:
\begin{equation}
\label{eq:residuals-delta}
    \bm{h}^{(\ell)}_i = \bar{\bm{h}}^{(\ell)} + \bm\delta_i.
\end{equation}
Given that cosine similarity utilizes only $\bar{\bm{h}}^{(\ell)}$, it ignores the set of deviations of $\bm\delta_i$, limiting to measure the difference that is formed by $\bm{\delta}_i$.

The representational structure distance computes the difference between the two spanning trees that are based on the L2 distance $\|\bm{h}^{(\ell)}_i - \bm{h}^{(\ell)}_j\|$ (see Equation~\ref{eq:adjacency}).
By applying the decomposition of representations of Equation~\ref{eq:residuals-delta}, the L2 distance can be written as $\|\bm{\delta}_i - \bm{\delta}_j\|$.
That is, the structural distance reflects the internal relational differences, whereas the cosine similarity measure abstracts differences via mean pooling.

In light of this difference between cosine similarity and structural distance, the TED results of multilingual models shown in Figures~\ref{fig:ted_eng_tgt} provide distinct differences among languages, while the cosine similarity ones shown in Figure~\ref{fig:cos_sim} exhibit almost no differences among languages.
This difference indicates that the abstract representations are aligned with English and that the internal relational differences depend on the language in multilingual models.

\subsection{Resources and Linguistic Features}

\paragraph{Resource influence.}
Multilingual models exhibit a TED cluster comprising high- and mid-resource languages.
Recall that the languages used in this study are classified into two clusters. high- and mid-resource languages (i.e., German, Japanese, Chinese, and Indonesian) and low-resource languages (i.e., Aranese, Guarani, and Jinghpaw), the cluster can be attributed to the high-resource languages.
Furthermore, the distinct difference between English and other languages in Figure~\ref{fig:ted_heatmaps} supports this finding, since Olmo3 7B Base is a monolingual model trained on English data.

\paragraph{Linguistic feature influence.}
Among the three low-resource languages in the multilingual models' results, Figures~\ref{fig:ted_eng_tgt}~and~\ref{fig:ted_zh_tgt} show that Aranese exhibits the lowest TED, while Jinghpaw exhibits the highest TED.
Given that Aranese is closer to English than other low-resource languages, linguistic similarity influences TED for low-resource languages.

\subsection{How Do LLMs Represent Diverse Languages?}
The cosine similarity results in Figure~\ref{fig:cos_sim} and TED results in Figure~\ref{fig:ted_eng_tgt} indicate that LLMs organize input sequences according to the language through each Transformer block and align abstract semantic representations across languages.
Multilingual LLMs exhibit relatively similar structural patterns for high- and mid-resource languages.
For unfamiliar, low-resource languages, the linguistically closer languages to English are easier to structurally align with English in representation space.

\subsection{How Does Language-Specific Post-Training Alter the Representations?}

Figure~\ref{fig:ted_post_train} shows that Qwen3 Swallow 8B CPT, a Japanese-specific post-trained model, exhibits divergence in TED from the base model within Japanese.
Conversely, the representational structure distance between English and other languages, including Japanese, remains invariant (see Figure~\ref{fig:ted_eng_tgt_post_train}).
These findings suggest that language-specific post-training modifies target language structures while preserving inter-lingual structural distances, allowing performance improvement in that language.

\section{Conclusion}
In this study, we investigate the multilinguality of LLMs from a representational structure perspective.
Our experiments indicate that the structural distance between languages depends on the models' familiarity with those languages, and that language-specific post-training refines the internal structure of that language while preserving its relationships to other languages.
Our findings highlight that LLMs distinguish familiar and unfamiliar languages in their internal structures and, post-training, retain the relationship between languages.

\clearpage
\section*{Limitations}
\paragraph{Language selection.}
We use a set of languages: English, German, Chinese, Japanese, Aranese, Guarani, and Jinghpaw.
Although this set of languages may limit the generality of our work, this set includes high-, mid-, and low-resource languages, multiple scripts, and typological diversity, allowing us to investigate multilinguality of LLMs from resource, scripts, and typological aspects.
This language set enables us to find that multilingual LLMs exhibit high similarity in representational structure across high- and low-resource languages despite their diverse typological features, indicating that this set is sufficient for this work's scope.

\paragraph{Scope of representational structures.}
We compare representational structures across languages and models.
The tree structures are model-internal summaries of inter-token relationships in LLM representation space, not linguistic parse trees.
Therefore, we interpret our findings as evidence about how LLMs organize token representations, and leave direct comparison with linguistic structures to future work.

\paragraph{Training data availability.}
As of writing, the multilingual models used in this study have not released their training data.
This limits our investigation into the influence of multilingual training data on the structural distance.
Testing multilingual models trained on controlled multilingual text can provide further insights beyond our work; this is a future direction for this work.
Our primary goal is to investigate how LLMs that exhibit advanced performance represent and process multiple languages, and this work provides insights into this.

\paragraph{Applications.}
Although we believe that our findings are helpful for applications, e.g., language adaptation, this study aims to reveal how LLMs process multiple languages from a structural perspective that captures holistic aspects, following prior efforts on the multilinguality of LLMs.

\section*{Ethical Considerations}
\paragraph{Reproducibility.}
We provide experimental settings, including software and computational resources, in Section~\ref{sec:experimental-settings} and Appendix~\ref{sec:detail-experimental-settings}.

\paragraph{Use of AI assistants.}
We use LLMs for a coding assistant and proofreading of the manuscript, and all the edits of code and paper are verified by the authors.

\paragraph{Data content.}
Although we use FLORES+, a widely used dataset, and its corresponding dataset, we have not assessed whether either dataset contains personally identifiable or offensive content, as they contain multiple languages, including low-resource languages.

\paragraph{License.}
We follow the license of the models, datasets, and software used in this study.

\section*{Acknowledgments}
This work was supported by JST BOOST, Japan Grant Number JPMJBS2423.

\bibliography{custom}

\clearpage
\appendix

\section{Experimental Details}
\label{sec:detail-experimental-settings}

\subsection{Models and Datasets}
Table~\ref{tab:model-hf} is the mapping table of model names used in our experiments and their HuggingFace IDs.
We obtain FLORES+ data from HuggingFace\footnote{\url{https://hf.co/datasets/openlanguagedata/flores_plus}} and corresponding Jinghpaw data from \citet{taguchi-etal-2025-languages}\footnote{\url{https://github.com/ctaguchi/LSLB}}.
We provide the licenses for these artifacts in Table~\ref{tab:license}.

\subsection{Implementations and Computational Resources}
In our experiments, we use HuggingFace Transformers~\cite{wolf-etal-2020-transformers} to load models and datasets.
We utilize \textsc{Sacre}BLEU~\cite{post-2018-call} and its official implementation\footnote{\url{https://github.com/mjpost/sacrebleu}} to compute BLEU, chrF, and TER.
To run \textsc{StructLens}, we use the official implementations\footnote{\url{https://github.com/naist-nlp/structlens}}.
For the AP-TED algorithm, we utilize the Python implementation\footnote{\url{https://github.com/JoaoFelipe/apted}}.
We use a single NVIDIA GeForce RTX 3090 GPU for the experiments involving the models with less than 8B, a single NVIDIA 6000 Ada GPU for experiments involving the Gemma3 12B PT and Qwen3 14B Base, and a single NVIDIA H100 GPU for Olmo3 32B Base.

\subsection{Prompt for Machine Translation}
We use the following prompt for machine translation experiments.

\begin{tcolorbox}[boxrule=1pt]
Translate the following \{source language\} text into \{target language\}. Output only the translation:

\{Text\}
\\
Translation: 

\end{tcolorbox}

\subsection{Non-Structural Metrics}
\label{sec:detail-non-structural-metrics-settings}

\paragraph{L2 distance.}
We also measure L2 distance to investigate whether norm information is meaningful for representation analysis.
The L2 distance between source and target languages is provided as:
\begin{equation}
    \|\bar{\bm{h}}^{(\ell)}_{src} - \bar{\bm{h}}^{(\ell)}_{tgt}\|.
\end{equation}

\begin{table}[t]
    \centering
    \resizebox{0.9\linewidth}{!}{
    \begin{tabular}{ll}
        \toprule
           Model  & HuggingFace ID \\
           \midrule
           Gemma3 4B PT & google/gemma-3-4b-pt \\
           Gemma3 12B PT & google/gemma-3-12b-pt \\
           Qwen3 8B Base  & Qwen/Qwen3-8B-Base \\
           Qwen3 14B Base  & Qwen/Qwen3-14B-Base \\
           Qwen3 8B & Qwen/Qwen3-8B \\
           Olmo3 7B Base & allenai/Olmo-3-1025-7B \\
           Olmo3 32B Base & allenai/Olmo-3-1125-32B \\
           Qwen3 Swallow 8B CPT & tokyotech-llm/Qwen3-Swallow-8B-CPT-v0.2 \\
        \bottomrule
    \end{tabular}
    }
    \caption{Model names and HuggingFace IDs.}
    \label{tab:model-hf}
\end{table}

\begin{table}[t]
    \centering
    \resizebox{0.6\linewidth}{!}{
    \begin{tabular}{ll}
        \toprule
           Artifact  & License \\
           \midrule
           Gemma3 4B PT & Gemma \\
           Gemma3 12B PT & Gemma \\
           Qwen3 8B Base  & Apache 2.0 \\
           Qwen3 14B Base  & Apache 2.0 \\
           Qwen3 8B & Apache 2.0  \\
           Olmo3 7B Base & Apache 2.0 \\
           Olmo3 32B Base & Apache 2.0 \\
           Qwen3 Swallow 8B CPT & Apache 2.0 \\
           FLORES+ & CC BY-SA 4.0 \\
           FLORES+ (Jinghpaw) & CC-BY-SA-NC \\
        \bottomrule
    \end{tabular}
    }
    \caption{Artifacts license.}
    \label{tab:license}
\end{table}

\section{Detailed Experimental Results}
\label{sec:detail-experimental-results}

\subsection{Non-Structural Metrics}
\label{sec:detail-non-structural-metrics}

\paragraph{Token length.}
We also measure token length, which is provided in Table~\ref{tab:token_length}, showing that low-resource languages exhibit larger token length than other languages in multilingual models.

\begin{table*}[t]
  \centering
  \resizebox{\textwidth}{!}{
  \begin{tabular}{lcccccccc}
    \toprule
    Model & \multicolumn{1}{c}{English} & \multicolumn{1}{c}{German} & \multicolumn{1}{c}{Japanese} & \multicolumn{1}{c}{Chinese} & \multicolumn{1}{c}{Indonesian} & \multicolumn{1}{c}{Aranese} & \multicolumn{1}{c}{Guarani} & \multicolumn{1}{c}{Jinghpaw} \\
    \midrule
    Gemma3 4B PT & $26.96_{\pm 8.77}$ & $35.12_{\pm 11.76}$ & $32.31_{\pm 11.41}$ & $29.38_{\pm 11.89}$ & $30.59_{\pm 10.12}$ & $45.91_{\pm 15.31}$ & $51.13_{\pm 17.70}$ & $54.52_{\pm 18.92}$ \\
    Qwen3 8B Base & $26.33_{\pm 8.94}$ & $40.79_{\pm 14.19}$ & $37.03_{\pm 13.57}$ & $26.98_{\pm 11.79}$ & $40.59_{\pm 13.58}$ & $48.78_{\pm 16.51}$ & $54.96_{\pm 19.28}$ & $60.07_{\pm 21.13}$ \\
    Olmo3 7B Base & $25.88_{\pm 8.51}$ & $40.74_{\pm 14.06}$ & $58.64_{\pm 20.30}$ & $48.90_{\pm 18.88}$ & $40.13_{\pm 13.31}$ & $48.68_{\pm 16.49}$ & $56.34_{\pm 19.91}$ & $59.79_{\pm 21.06}$ \\
    \bottomrule
  \end{tabular}
  }
  \caption{Token length statistics across models and subsets. Subscripts denote standard deviation.}
  \label{tab:token_length}
\end{table*}

\paragraph{MT results (detail).}
Table~\ref{tab:mt} shows the MT performance for each metric.
The models exhibit a better score for Aranese among low-resource languages, which suggests that linguistic similarity to English influences the MT performance.

\begin{table*}[t]
  \centering
  \resizebox{\textwidth}{!}{
  \begin{tabular}{lccccccccccccccccccccc}
    \toprule
    Model & \multicolumn{3}{c}{German $\rightarrow$ English} & \multicolumn{3}{c}{Japanese $\rightarrow$ English} & \multicolumn{3}{c}{Chinese $\rightarrow$ English} & \multicolumn{3}{c}{Indonesian $\rightarrow$ English} & \multicolumn{3}{c}{Aranese $\rightarrow$ English} & \multicolumn{3}{c}{Guarani $\rightarrow$ English} & \multicolumn{3}{c}{Jinghpaw $\rightarrow$ English} \\
    & BLEU~$\uparrow$ & CHRF~$\uparrow$ & TER~$\downarrow$ & BLEU~$\uparrow$ & CHRF~$\uparrow$ & TER~$\downarrow$ & BLEU~$\uparrow$ & CHRF~$\uparrow$ & TER~$\downarrow$ & BLEU~$\uparrow$ & CHRF~$\uparrow$ & TER~$\downarrow$ & BLEU~$\uparrow$ & CHRF~$\uparrow$ & TER~$\downarrow$ & BLEU~$\uparrow$ & CHRF~$\uparrow$ & TER~$\downarrow$ & BLEU~$\uparrow$ & CHRF~$\uparrow$ & TER~$\downarrow$ \\
    \midrule
    Gemma3 4B PT & $35.36$ & $60.42$ & $53.79$ & $21.31$ & $48.85$ & $73.37$ & $22.82$ & $50.14$ & $70.32$ & $33.80$ & $59.47$ & $57.21$ & $16.39$ & $44.15$ & $84.00$ & $3.88$ & $23.36$ & $146.31$ & $1.22$ & $15.46$ & $234.59$ \\
    Qwen3 8B Base & $36.90$ & $62.09$ & $51.04$ & $22.25$ & $51.80$ & $69.50$ & $24.99$ & $53.76$ & $66.24$ & $36.07$ & $62.56$ & $51.09$ & $19.48$ & $48.83$ & $76.56$ & $5.41$ & $27.20$ & $103.62$ & $3.60$ & $23.08$ & $118.63$ \\
    Olmo3 7B Base & $35.42$ & $61.17$ & $53.38$ & $19.44$ & $48.11$ & $75.48$ & $21.87$ & $49.91$ & $71.76$ & $31.70$ & $57.94$ & $57.19$ & $13.12$ & $41.95$ & $91.02$ & $2.40$ & $22.23$ & $181.67$ & $1.99$ & $17.28$ & $173.27$ \\
    \bottomrule
  \end{tabular}
  }
  \caption{MT evaluation results. Higher is better ($\uparrow$) for BLEU/CHRF, lower is better ($\downarrow$) for TER.}
  \label{tab:mt}
\end{table*}

\begin{figure*}[t]
    \centering
    \includegraphics[width=0.6\linewidth]{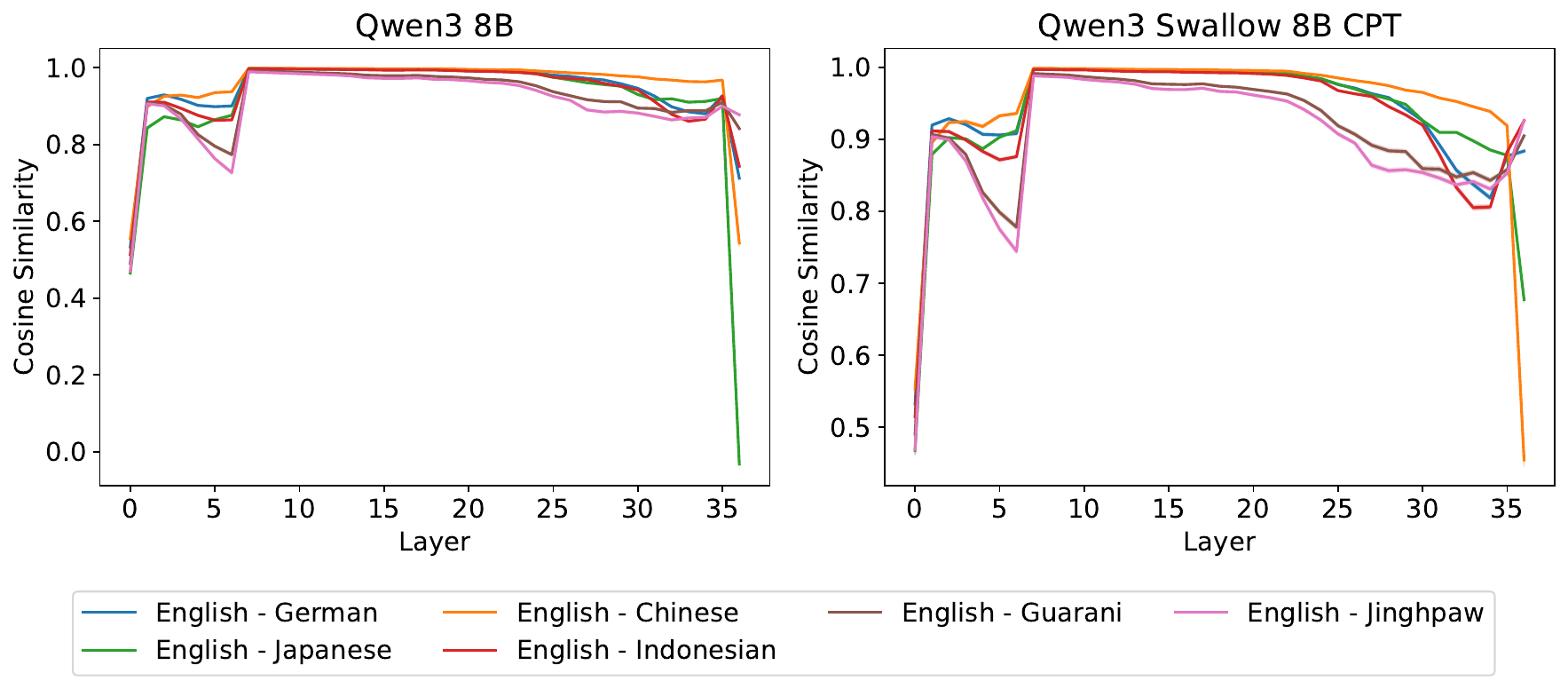}
    \caption{Cosine similarity on post-trained models.}
    \label{fig:cos_sim_post_trained}
\end{figure*}

\paragraph{Cosine similarity for post-trained models.}
Figure~\ref{fig:cos_sim_post_trained} post-training results indicate that post-training has less influence on cosine similarity.

\paragraph{L2 distance.}

\begin{figure*}[t]
    \centering
    \includegraphics[width=\linewidth]{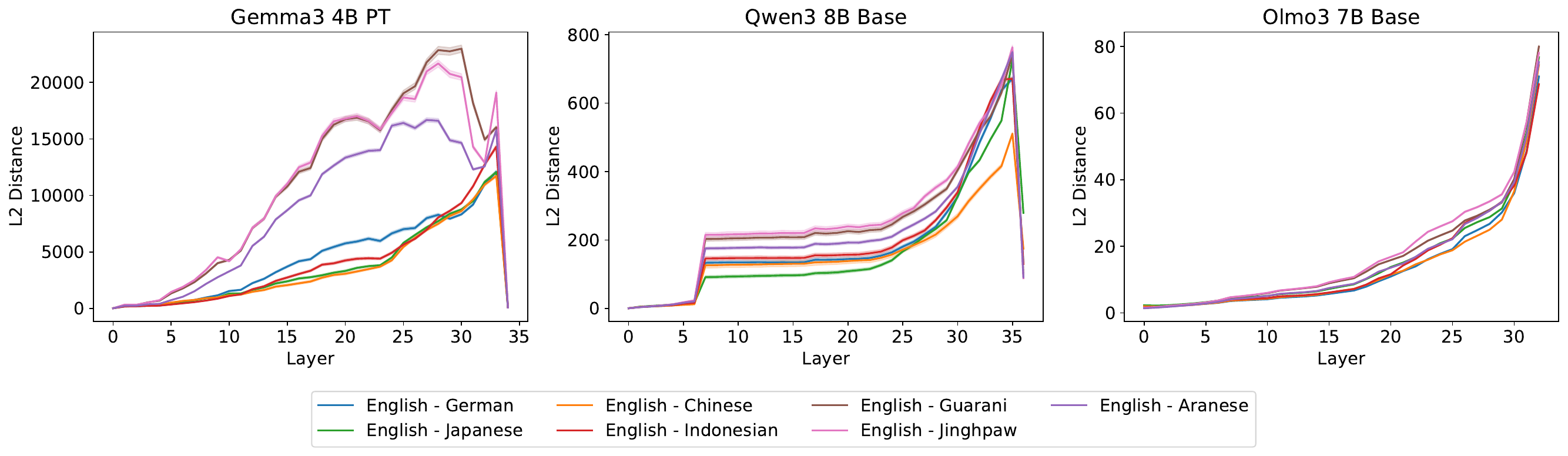}
    \caption{L2 distance on pre-trained models}
    \label{fig:l2_dist}
\end{figure*}

\begin{figure*}[t]
    \centering
    \includegraphics[width=0.6\linewidth]{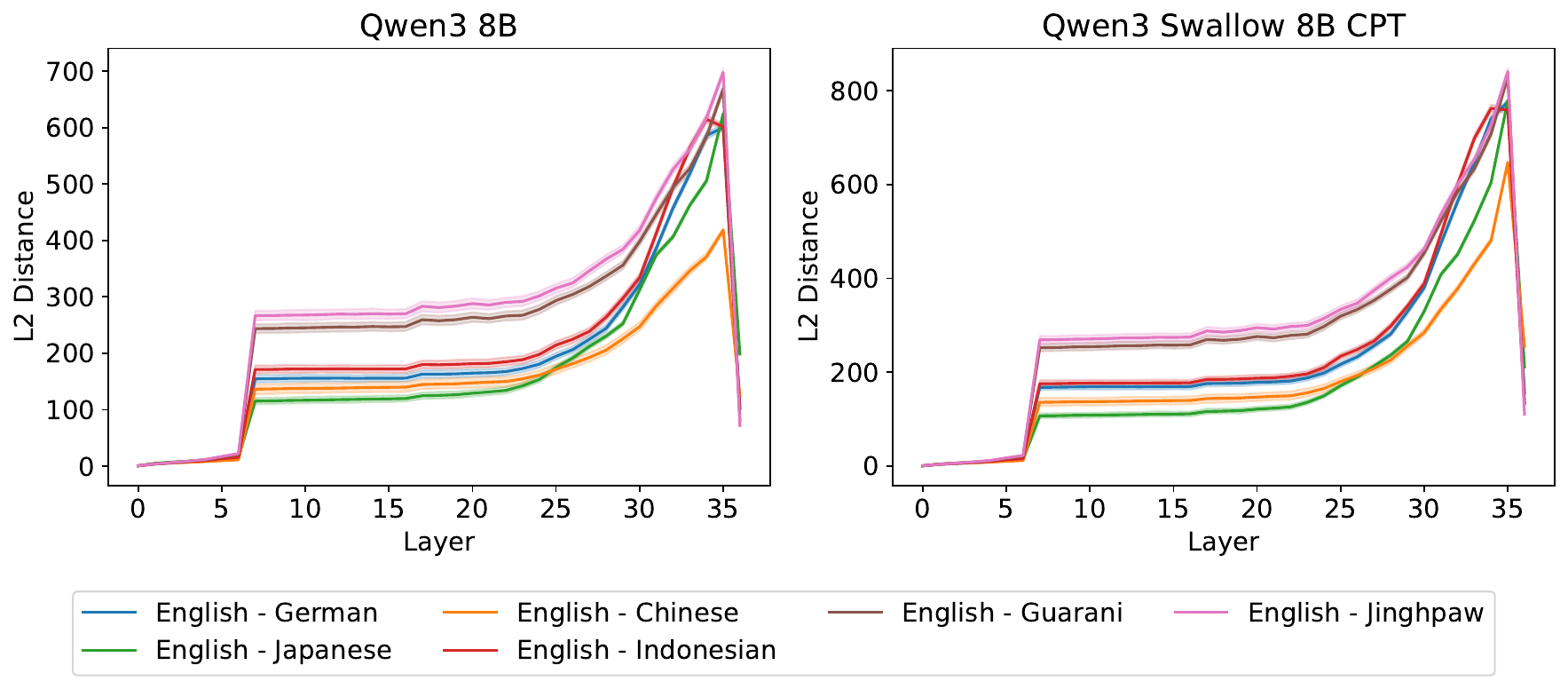}
    \caption{L2 distance on post-trained models}
    \label{fig:l2_dist_post_trained}
\end{figure*}

Figures~\ref{fig:l2_dist}~and~\ref{fig:l2_dist_post_trained} show L2 distance for pre- and post- trained models, respectively.
Given the nature of residual streams, the monotonic increments of L2 distance are obvious.

\subsection{Structural Metric}
\label{appendix:structural-distance}

\begin{figure*}
\begin{subfigure}[b]{0.48\linewidth}
    \centering
    \includegraphics[width=\linewidth]{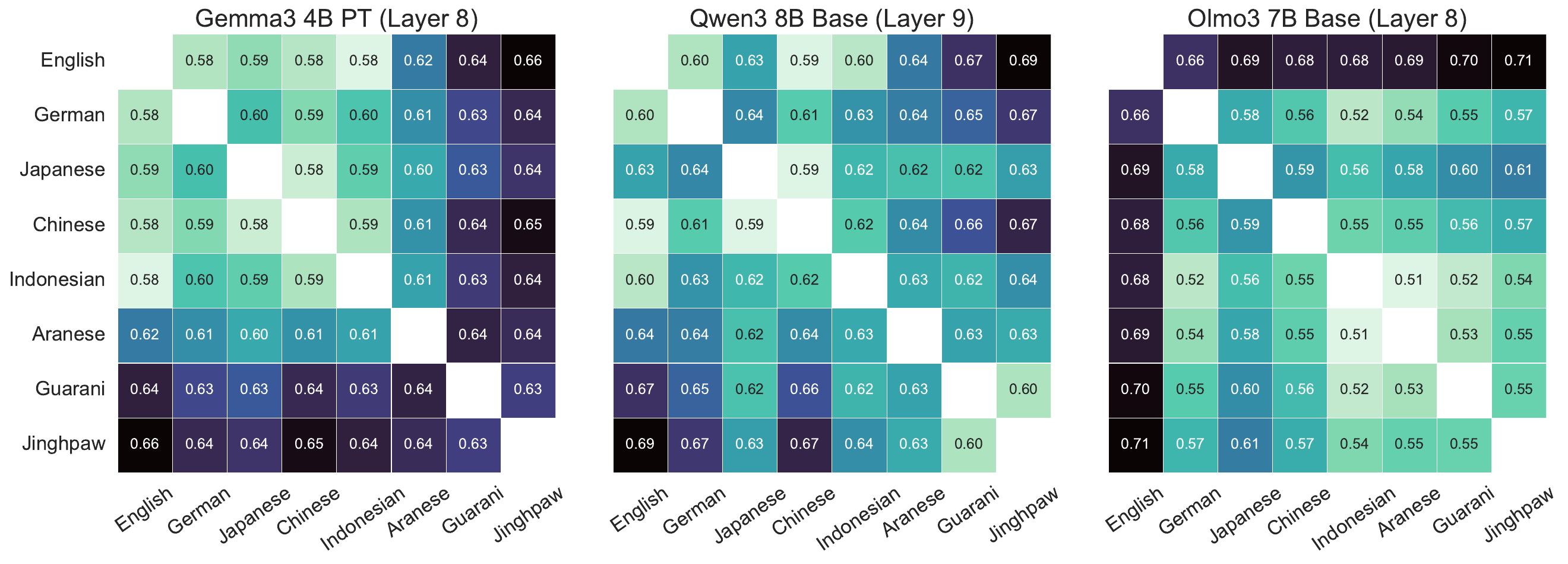}
    \caption{Low layers.}
    \label{fig:heatmaps_ted_low}
\end{subfigure}
\begin{subfigure}[b]{0.48\linewidth}
    \centering
    \includegraphics[width=\linewidth]{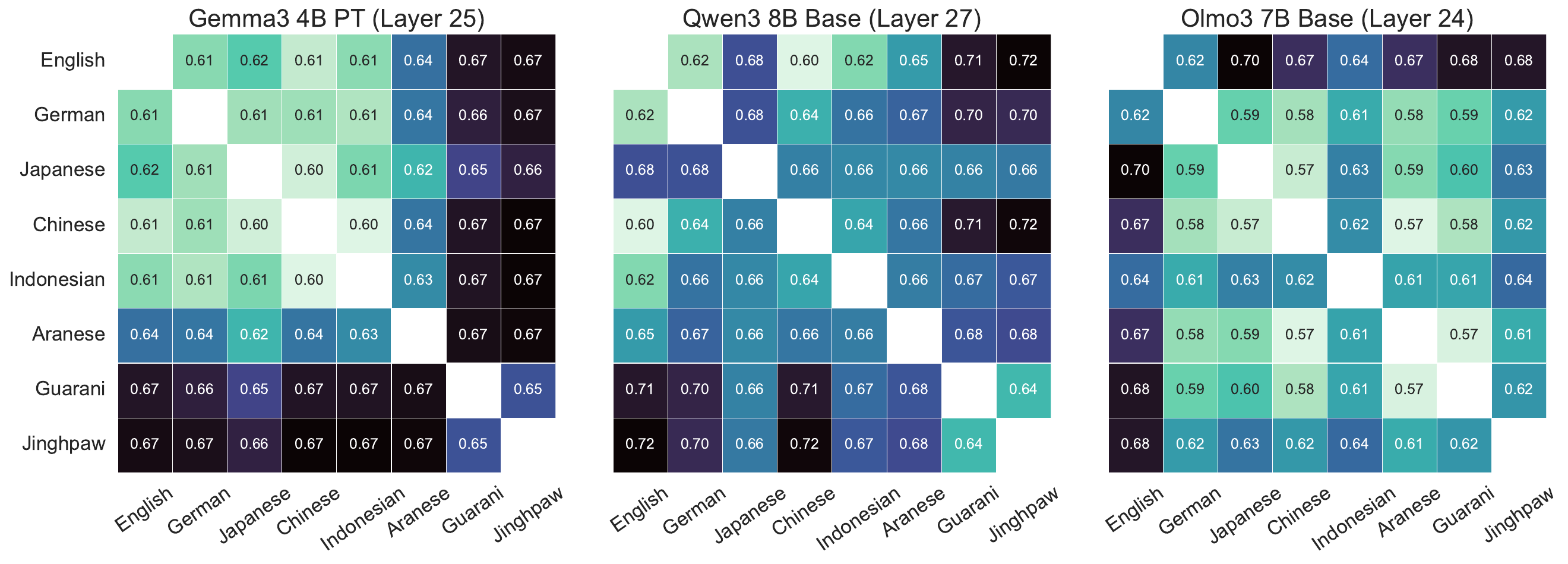}
    \caption{High layers.}
    \label{fig:heatmaps_ted_high}
\end{subfigure}
\caption{Heatmaps of TED for each language pair.}
\label{fig:heatmaps_ted_small}
\end{figure*}

\begin{figure*}
\begin{subfigure}[b]{0.48\linewidth}
    \centering
    \includegraphics[width=\linewidth]{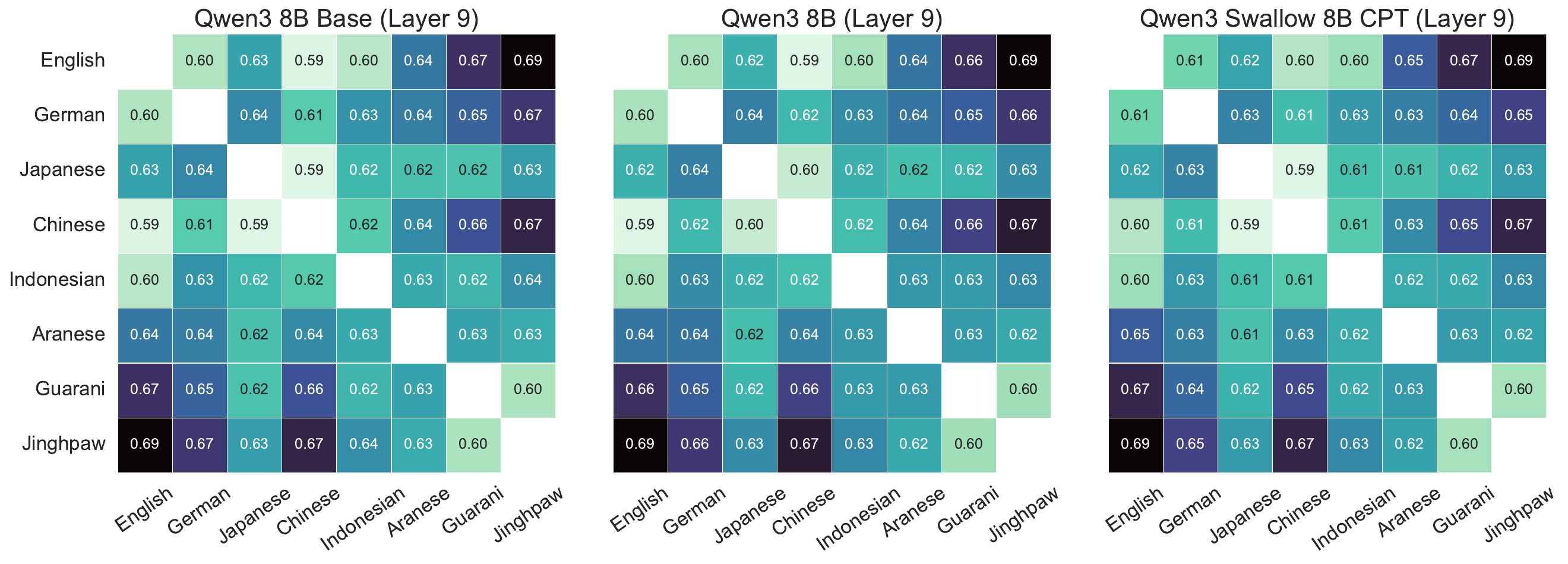}
    \caption{Low layers.}
    \label{fig:heatmaps_ted_low_post}
\end{subfigure}
\begin{subfigure}[b]{0.48\linewidth}
    \centering
    \includegraphics[width=\linewidth]{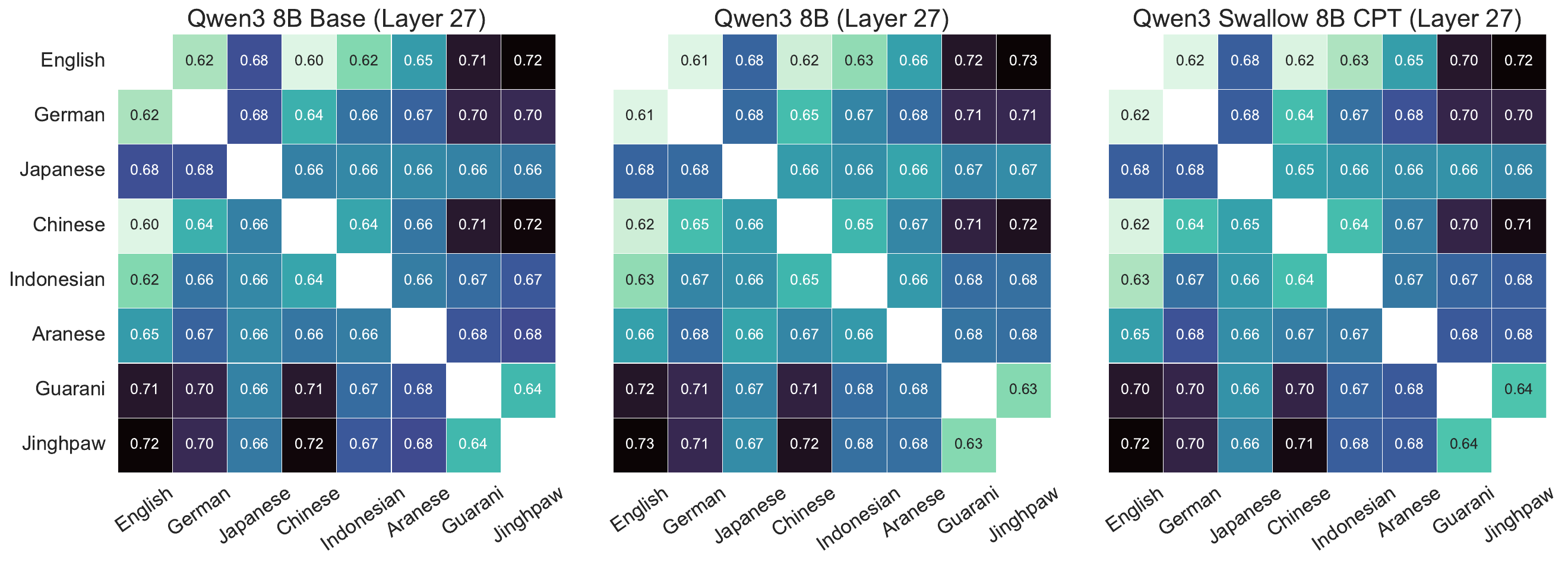}
    \caption{High layers.}
    \label{fig:heatmaps_ted_high_post}
\end{subfigure}
\caption{Heatmaps of TED in pre- and post-trained models.}
\label{fig:heatmaps_ted_post}
\end{figure*}

\begin{figure*}
\centering
\begin{subfigure}[b]{0.48\linewidth}
    \centering
    \includegraphics[width=\linewidth]{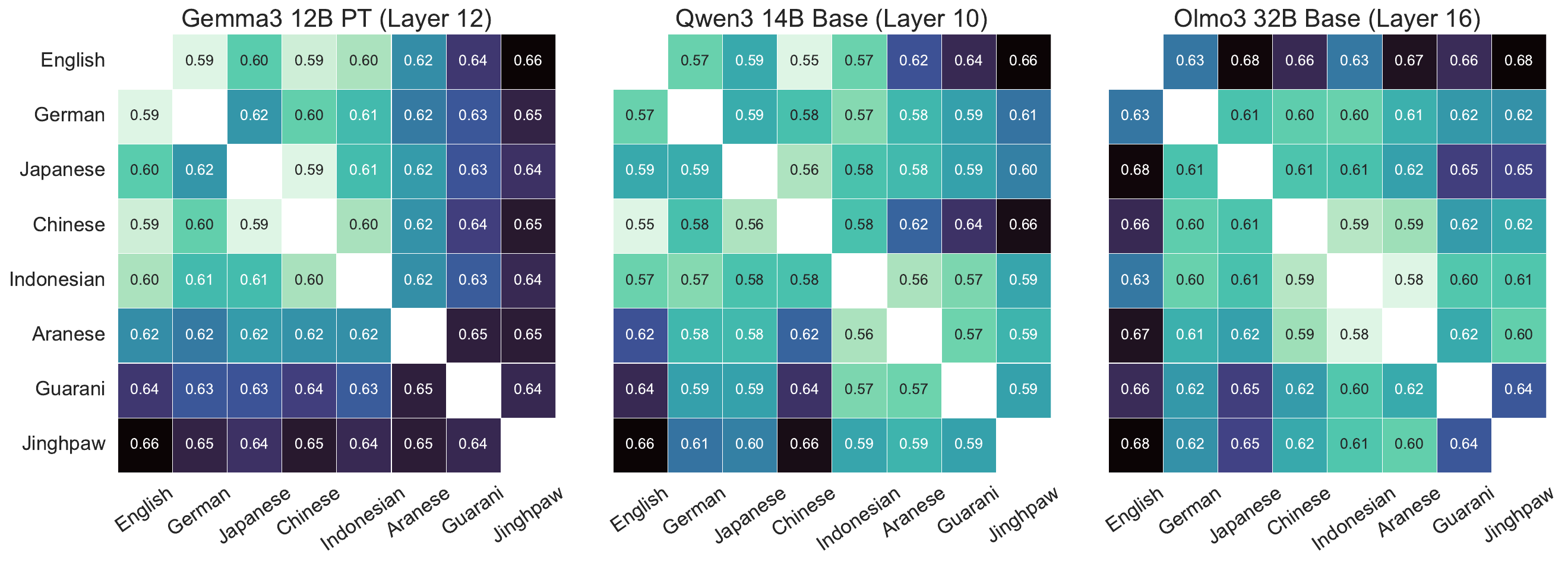}
    \caption{Low layers.}
    \label{fig:heatmaps_ted_low_large}
\end{subfigure}
\begin{subfigure}[b]{0.48\linewidth}
    \centering
    \includegraphics[width=\linewidth]{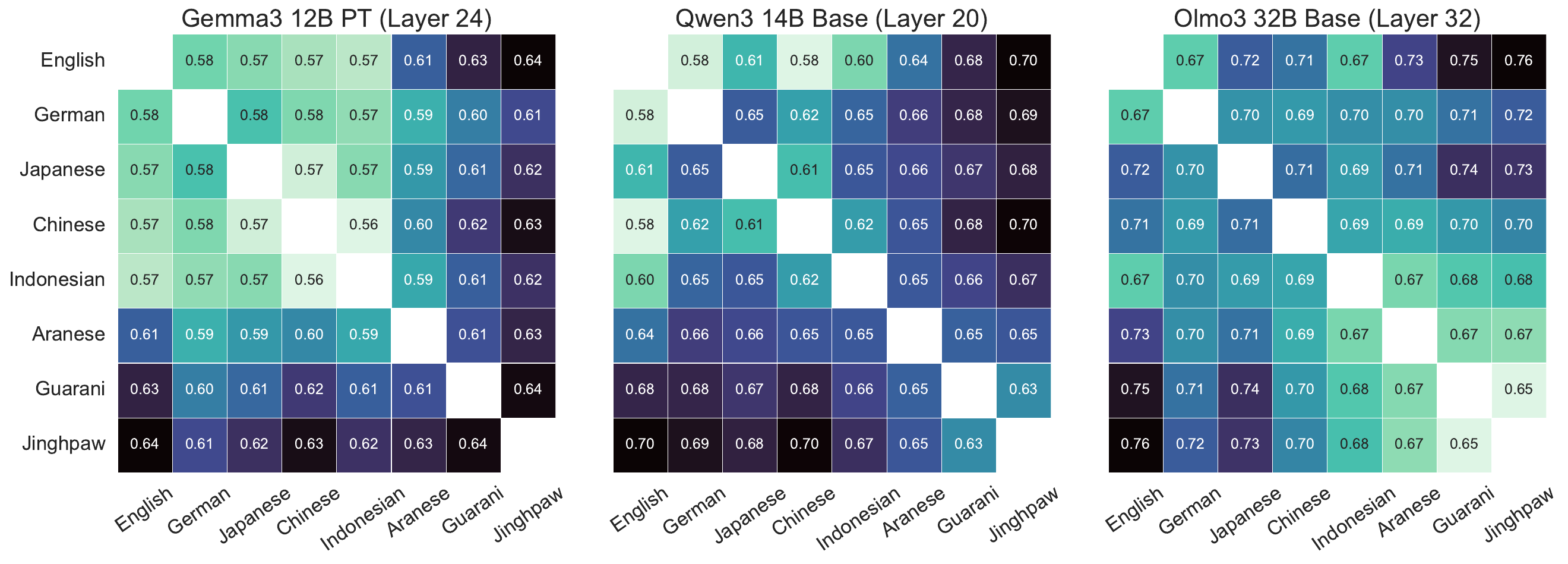}
    \caption{Middle layers.}
    \label{fig:heatmaps_ted_middle_large}
\end{subfigure}
\begin{subfigure}[b]{0.48\linewidth}
    \centering
    \includegraphics[width=\linewidth]{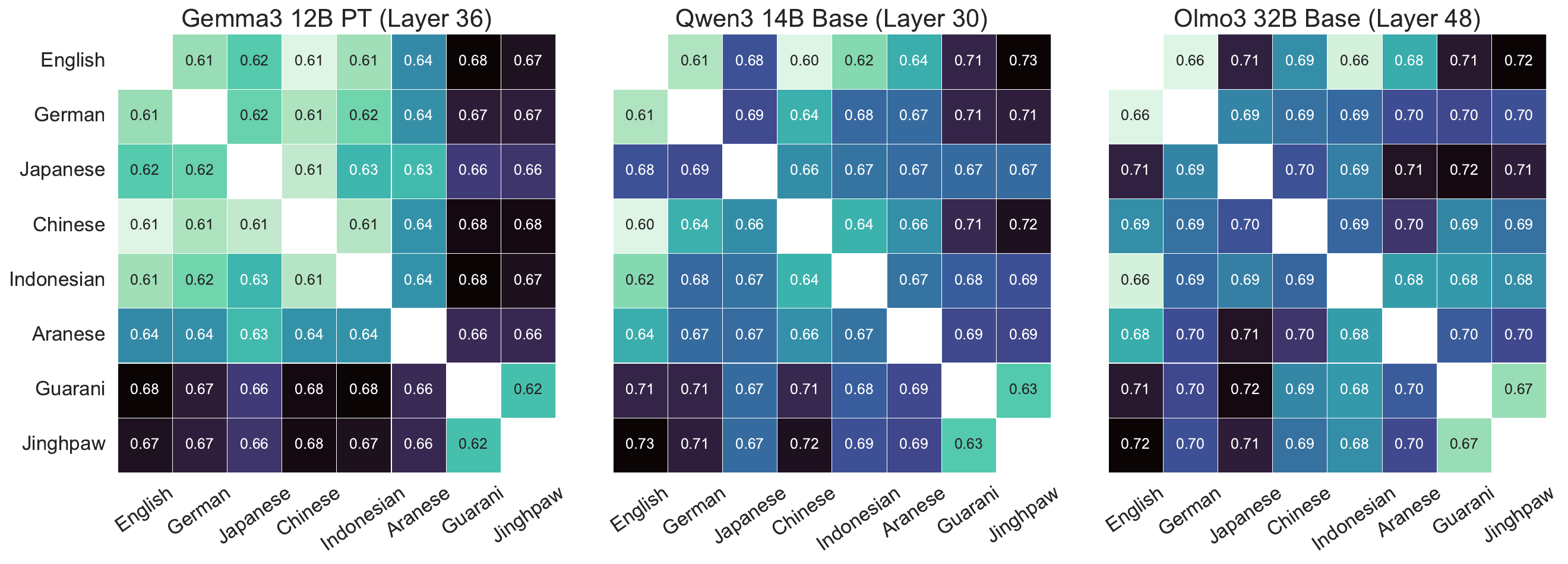}
    \caption{High layers.}
    \label{fig:heatmaps_ted_high_large}
\end{subfigure}
\caption{Heatmaps of TED in larger models.}
\label{fig:heatmaps_ted_large}
\end{figure*}

We visualize TED for each language pair in low and high layers of pre- and post-trained models in Figures~\ref{fig:heatmaps_ted_small}~and~\ref{fig:heatmaps_ted_post}, and that in low, middle, and high layers in Figure~\ref{fig:heatmaps_ted_large}.
While smaller models exhibit consistent structural similarity patterns across layers, the monolingual larger model shows different patterns, where distinct English patterns shown in the smaller model disappear in the middle and high layers.

\end{document}